\documentclass{article}

\usepackage{times}
\usepackage{graphicx} 
\usepackage{subfigure} 

\usepackage{natbib}

\usepackage{algorithm}
\usepackage{algorithmic}

\usepackage{hyperref}

\usepackage[accepted]{icml2017}

\usepackage{amsmath}
\usepackage{amsfonts}
\usepackage{glossaries}

\icmltitlerunning{Decoupled Learning of Environment Characteristics for Safe Exploration}

\newacronym{mdp}{MDP}{Markov Decision Proces}
\newacronym{dqn}{DQN}{deep Q-network}
\newacronym{dnn}{DNN}{deep neural network}
\newacronym{relu}{ReLUs}{rectified linear units}

\begin{document} 

\twocolumn[
\icmltitle{Decoupled Learning of Environment Characteristics for Safe Exploration}



\icmlsetsymbol{equal}{*}

\begin{icmlauthorlist}
\icmlauthor{Pieter Van Molle}{ugent}
\icmlauthor{Tim Verbelen}{ugent}
\icmlauthor{Steven Bohez}{ugent}
\icmlauthor{Sam Leroux}{ugent}
\icmlauthor{Pieter Simoens}{ugent}
\icmlauthor{Bart Dhoedt}{ugent}
\end{icmlauthorlist}

\icmlaffiliation{ugent}{Authors are with Ghent University - imec, IDLab, Department of Information Technology}

\icmlcorrespondingauthor{Pieter Van Molle}{pieter.vanmolle@ugent.be}

\icmlkeywords{reinforcement learning, transfer learning, safe exploration}

\vskip 0.3in
]



\printAffiliationsAndNotice{} 

\begin{abstract} 
Reinforcement learning is a proven technique for an agent to learn a task. However, when learning a task using reinforcement learning, the agent cannot distinguish the characteristics of the environment from those of the task. This makes it harder to transfer skills between tasks in the same environment. Furthermore, this does not reduce risk when training for a new task. In this paper, we introduce an approach to decouple the environment characteristics from the task-specific ones, allowing an agent to develop a sense of survival. We evaluate our approach in an environment where an agent must learn a sequence of collection tasks, and show that decoupled learning allows for a safer utilization of prior knowledge.
\end{abstract} 

\section{Introduction}
\label{introduction}

When using traditional reinforcement learning to train an agent for a specific task in an environment, the agent does not differentiate between the characteristics of the environment, and those of the task. This does not allow for an easy transfer of skills between tasks.

The above behavior can be problematic in real-world scenarios, where gathering experience is both costly and dangerous. Consider a warehouse for example, where autonomous drones are deployed. During training of the drones' policy, many have been lost due to crashes \cite{gandhi2017learning}. Having to lose another series of drones when the objective changes would be far from optimal.

A better approach would be for the drones to have a notion of safety, or survival skills, regardless of their current task. In this paper, we introduce an approach to learn these survival skills independent of a task, by decoupling the environment characteristics from the task-specific ones when learning said task. We show that an agent that retains these skills between tasks exhibits a safer behavior than an agent that does not.

\section{Reinforcement learning}
\label{reinforcement learning}

In the reinforcement learning framework \cite{sutton1998reinforcement}, an agent interacts with its environment in a sequence of observations, actions and rewards. At each time-step $t$, the agent follows its policy $\pi$ to take an action $a_t$, with respect to the observed state $s_t$ of the environment. This results in a reward $r_{t}$ and a new state $s_{t+1}$ for the next time-step.

The objective of the reinforcement learning framework is to find a policy that maximizes the expected discounted return $R_t$

\begin{equation}
R_t = r_t + \gamma r_{t+1} + \gamma^2 r_{t+2} + \ldots
\end{equation}

where $\gamma$ is a discount factor between 0 and 1. Formally, this means finding a policy that follows the optimal action-value function $Q^*(s, a)$, defined as the maximum expected discounted return for taking an action $a$, given an observed state $s$, and following the optimal policy onwards.

\begin{equation}
Q^*(s, a) = \max_\pi \mathbb{E} \left[ R_t | s_t = s, a_t = a, \pi \right]
\end{equation}

Q-learning \cite{watkins1992q} is an off-policy algorithm that iteratively learns the optimal action-value function, by executing the following update rule:

\begin{equation}
Q(s, a) \leftarrow Q(s, a) + \alpha (r + \gamma \max_{a'} Q(s', a') - Q(s, a)).
\end{equation}

Deep Q-learning \cite{mnih2013playing} approximates the optimal action-value function by using a \gls*{dqn}. This is a deep neural network, parameterized by $\theta$, that represents $Q(s, a; \theta)$.

\section{Decoupled learning}
\label{decoupled_learning}

When training for a task, an agent can gain a notion of survival by decoupling the environmental reward signals from the task-specific ones. This way, the agent can independently learn the environment characteristics. Formally, we define a reward function $R_\varepsilon(s)$ for the environment, and a reward function $R_i(s)$ for task $i$.

\begin{figure}[t]
\vskip 0.2in
\begin{center}
\centerline{\includegraphics[width=0.65\columnwidth]{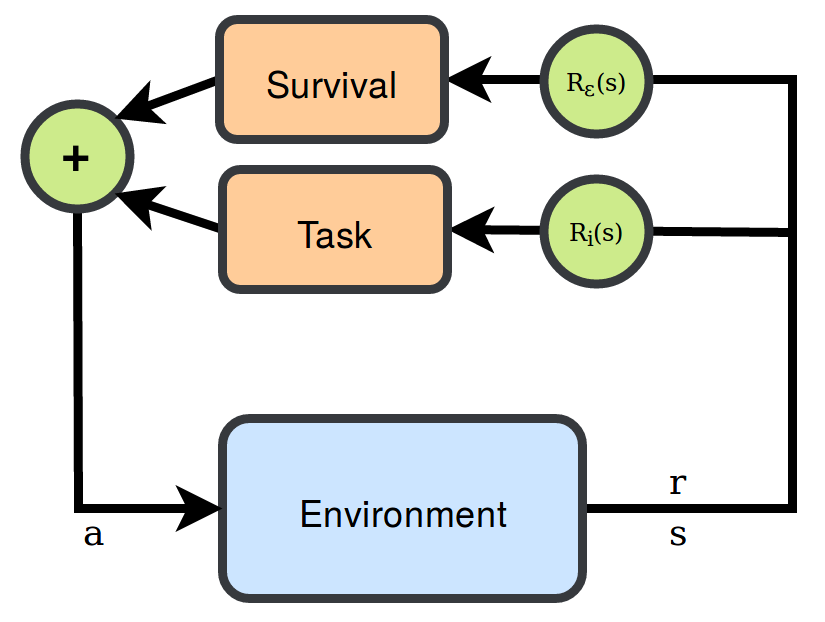}}
\caption{Architecture for decoupled learning.}
\label{fig:diagram}
\end{center}
\vskip -0.2in
\end{figure}

We can integrate decoupled learning in the Q-learning framework, by decomposing the action-value function

\begin{equation}
Q(s, a) = Q_\varepsilon(s, a) + Q_i(s, a)
\end{equation}

where $Q_\varepsilon(s, a)$ is defined as the action-value function for survival and $Q_i(s, a)$ as the action-value function for task $i$. We can learn both functions iteratively by applying the appropriate reward function during updates, instead of using the global reward. This is illustrated by Figure~\ref{fig:diagram}.

When training for a new task, we can leave the learned survival function $Q_\varepsilon(s, a)$ unmodified, and must only learn the new task function. Furthermore, the survival function is used to safely navigate the environment while gathering experience for the new task.

As an example, consider the traditional cliff walking problem \cite{sutton1998reinforcement}, where an agent is separated from its goal by a cliff. The agent receives a positive reward reaching its goal, and a negative reward when it falls into the cliff. In this problem, we can interpret a negative reward as an environmental punishment and a positive reward as completing the task. In doing so, the cliff walking problem is transformed into a part inherent to the environment (don't fall into the cliff), and a part specific for the task (reach the goal). When training an agent for this problem, we decouple the environment characteristics by propagating a negative reward to the survival function, and a positive reward to the task function.

\section{Experiments}
\label{experiments}

\begin{figure}[t]
\vskip 0.2in
\begin{center}
\centerline{\includegraphics[width=0.65\columnwidth]{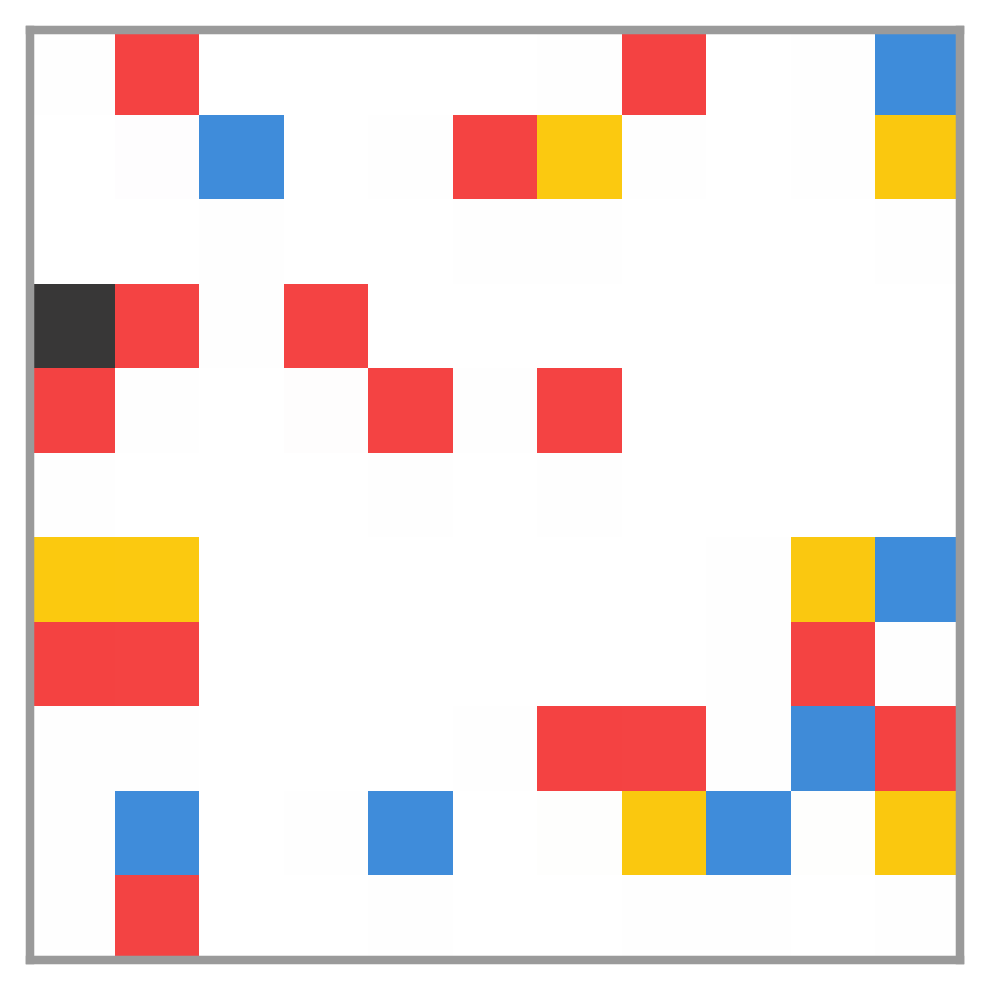}}
\caption{The environment. The agent (black) has to avoid the obstacles (red) and gather the collectibles (yellow or blue, depending on the task).}
\label{fig:setup}
\end{center}
\vskip -0.2in
\end{figure}

We evaluate our approach in an 11x11 gridworld environment, as seen on Figure~\ref{fig:setup}. In this environment, an agent must perform a collection task, while avoiding obstacles. At the beginning of an episode, the agent spawns on a random space. Each other space can spawn either a collectible or an obstacle. These spawn with a probability of 0.05 for the collectibles and 0.15 for the obstacles. There are two types of collectibles, distinguished by color. A task consists of gathering as many collectibles of a single type. When an agent grabs a collectible, a new one of the same type appears on a random empty space. If the taken collectible was of the desired type, the agent receives a +1 reward. When crashing into an obstacle, the agent receives a -1 reward, and the episode ends instantly. Otherwise, an episode lasts for a maximum of 50 steps.

We train an agent on the first collection task, change the task, and apply different methods to train the agent on the second task. We compare three methods:

\paragraph{Naive learning} To learn the first task, we use a single neural network to represent $Q(s, a; \theta)$. The weights $\theta$ of this network are randomly initialized. When learning the second task, we randomly re-initialize these weights.

\paragraph{Transfer learning} Once again, a single neural network with weights $\theta$ is used to learn the first task. Next, the trained weights are used to bootstrap learning the second task.

\paragraph{Decoupled learning} We use our decoupled learning approach, using two neural networks, to represent both the survival function $Q_\varepsilon(s, a; \theta)$ and the task-specific function $Q_i(s, i; \phi)$, during the first task. When training for the second task, we reuse the survival network, and only train a new network for the task.

The RGB state representation serves as input for each network. This input layer is followed by a convolutional layer with kernel size 3x3 and 64 filters, a max pooling layer, and two more convolutional layers with kernel size 3x3 and 32 filters. Another max pooling layer is followed by two fully connected layers, with 64 and 16 hidden units respectively. The output of each network consists of four values, representing the estimated Q-values for each of the four possible actions.

We use a replay memory of 10,000 experiences, which is continuously updated as the agent learns the first task. For the second task, a new replay memory is generated in different ways depending on the evaluated approach. In the naive case, the replay memory is initialized with random experiences. For transfer learning, the memory is filled with experiences of the agent performing the first task. When evaluating the decoupled approach, we use the survival network, trained during task one, to generate the replay memory.

We initialize the weights of each network using Xavier initialization \cite{glorot2010understanding}. Training is done for 60.000 episodes, by means of the Adam algorithm \cite{kingma2014adam}, with a minibatch size of 32 and a learning rate of 0.000025. For the second task, the naive agent is trained using $\varepsilon$-greedy learning, with $\varepsilon$ linearly annealed from 1.0 to 0.1. The transfer agent always follows its policy during training, and the decoupled agent uses an $\varepsilon$-greedy strategy, but instead of sampling over the entire action space, the agent picks an action from a subset of safe actions, provided by the survival network.

\begin{figure}[t]
\vskip 0.2in
\begin{center}
\centerline{\includegraphics[width=\columnwidth]{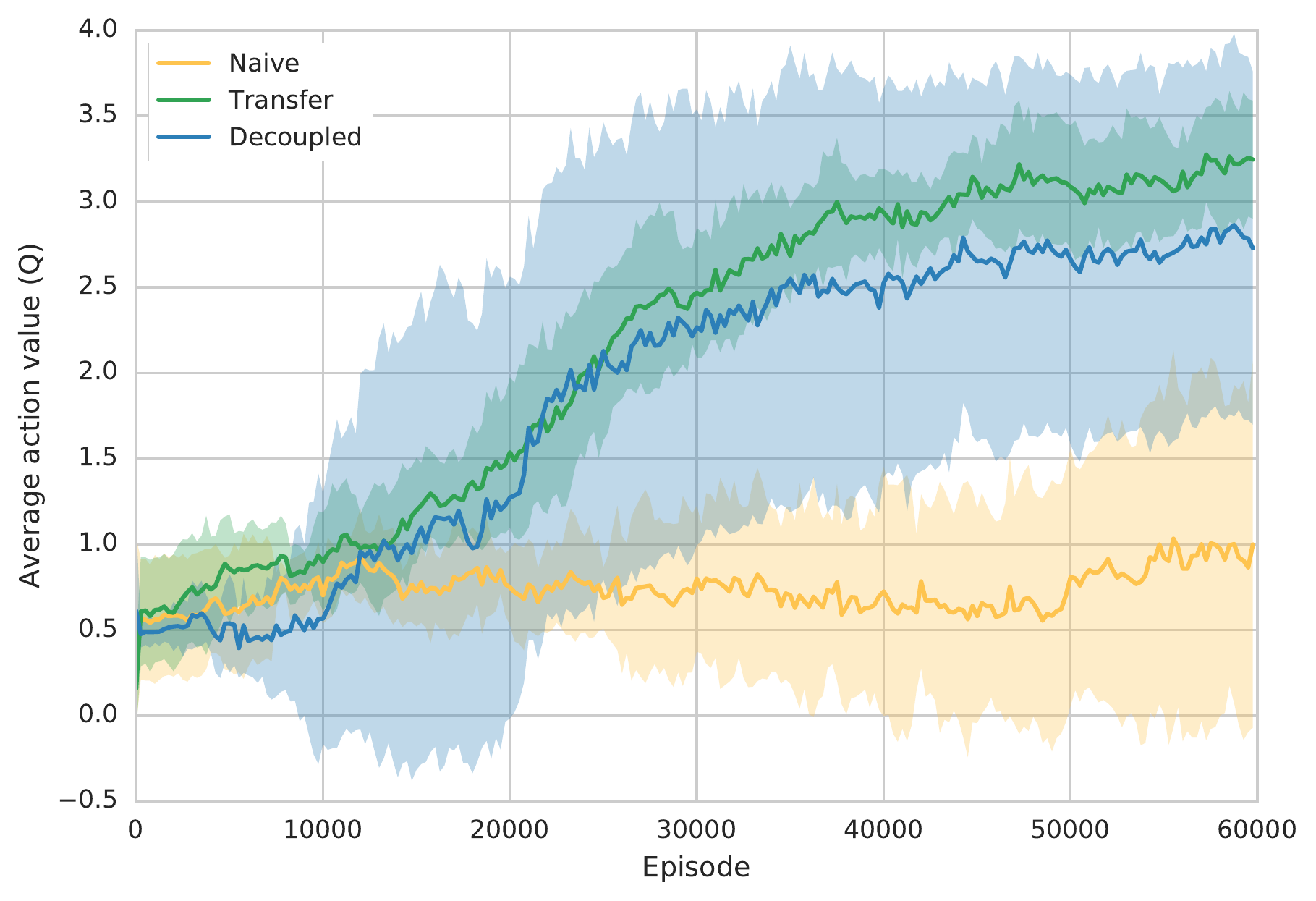}}
\caption{Learning curves for the second task, evaluated over a fixed set of states. All plots are an average over 9 random seeds.}
\label{fig:avg_q}
\end{center}
\vskip -0.2in
\end{figure}

To ascertain progress in re-training the agents, we apply each agent's action-value function as metric, as shown in Figure~\ref{fig:avg_q}. It shows how the decoupled agent equals the transfer agent in convergence rate, albeit with a higher variance. The naive agent fails to converge within the given time frame.

\begin{figure}[t]
\vskip 0.2in
\begin{center}
\centerline{\includegraphics[width=\columnwidth]{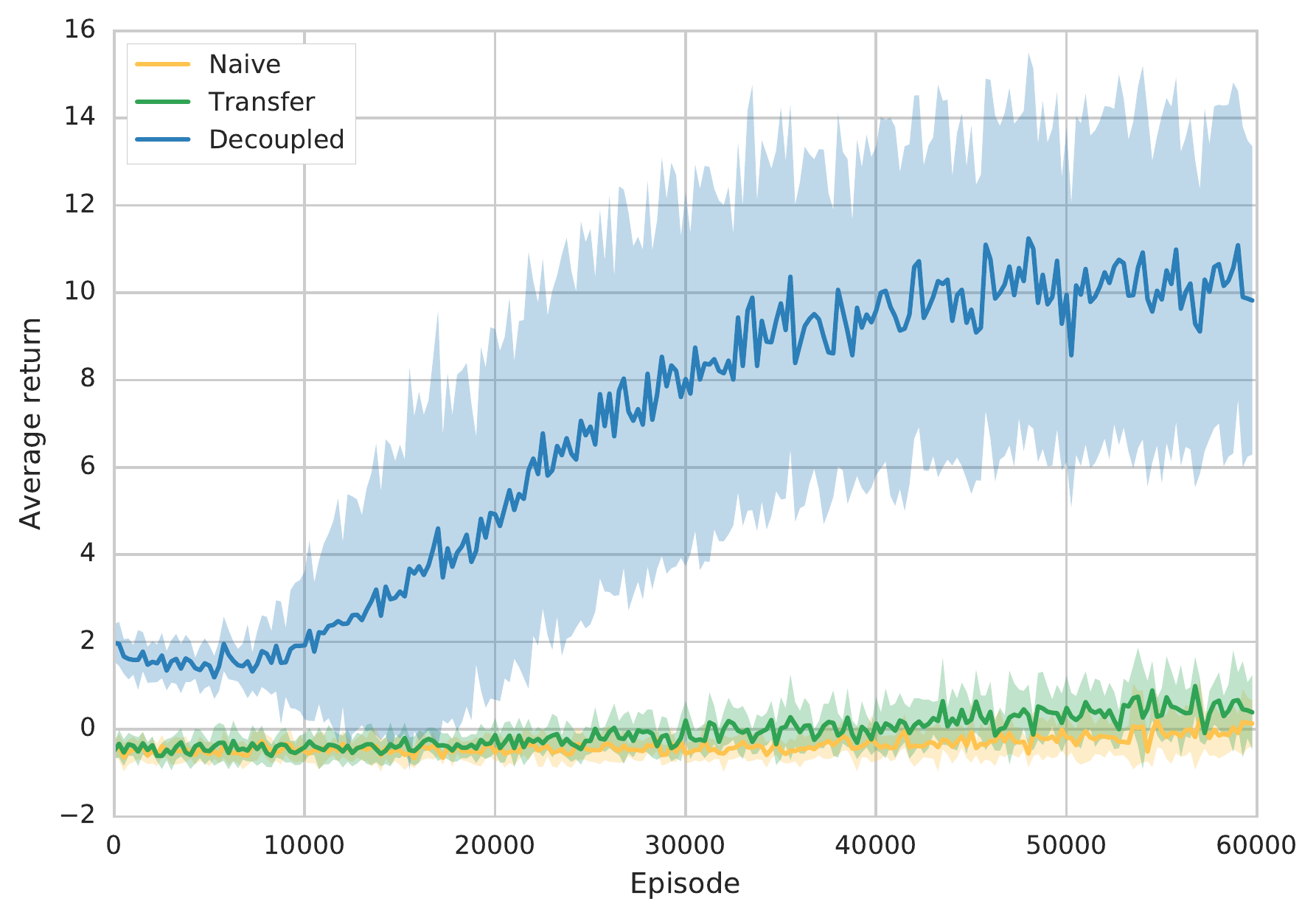}}
\caption{Comparison of the training performance of the naive, transfer and decoupled agent while learning the second task. All plots are an average over 9 random seeds.}
\label{fig:avg_return}
\end{center}
\vskip -0.2in
\end{figure}

Although the decoupled agent has the same convergence rate as the transfer agent, the first one learns a safer policy, resulting in a higher survivability. Because of this, the decoupled agent can take more steps each episode, which results in a higher episode return, as seen in Figure~\ref{fig:avg_return}.

\section{Related Work}
\label{related_work}

In transfer learning, knowledge from a source task is used to learn a target task better than if transfer learning were not used, according to some metric such as training time or total accumulated reward \cite{taylor2009transfer}. Different approaches exist to transfer knowledge between tasks. Autonomous shaping \cite{konidaris2006autonomous} tackles a sequence of goal-directed reinforcement learning tasks by separating each task in a problem-space representation, which can be different for each task, and an agent-space representation, which is the same across tasks. Using the latter representation, a shaping function is learned that provides value predictions for novel states across tasks as to speed up learning. The separation of problem-space and agent-space can also be extended to the level of options \cite{sutton1999between, konidaris2007building}. Transfer learning via inter-task mapping \cite{taylor2007transfer} uses hand coded task relationships to transform the action-value function from a source task to fit a target task with different state and/or action spaces. The MASTER method \cite{taylor2008autonomous} improves on the inter-task mapping by autonomously learning a mapping between a target task and one or more source tasks, by using experience the agent has gathered in the different task environments. When placing an agent in multiple environments, the agent itself is a common feature of each environment. By leveraging this stronger notion of an agent, the shared features framework \cite{konidaris2012transfer} allows for both transfer of knowledge between a source and a target task, and a way to learn portable skills through a sequence of tasks. The Actor-Mimic method \cite{parisotto2015actor} involves a single policy network learning to act in a set of distinct tasks through the guidance of an expert teacher for each task. Furthermore, the learned representation of the policy network enables generalizing to new tasks without expert guidance. The use of successor features \cite{barreto2016successor}, an extension of the successor representation \cite{dayan1993improving}, combined with a generalized framework for policy improvement, allows an agent to perform well on a novel task if it has seen a similar task before.

\section{Conclusion}
\label{conclusion}

In this paper, we present an approach for an agent to explicitly learn survival skills, by decoupling the environment characteristics from those of the task during training. This way, a learned representation of these characteristics can be transferred when training for a new task in the same environment. We compare our approach to both the naive method and the method of transfer learning. We evaluate each method by sequentially training an agent to gather different types of collectibles in a hostile environment. Our approach equals the method of transfer learning in terms of convergence, and, in addition, allows for a much safer utilization of prior knowledge, resulting in a higher episode return on average.

Following this paper, we plan to evaluate our approach in a real-world scenario, where a robot has to complete a series of tasks while crashing as little as possible.

\section*{Acknowledgements}
Steven Bohez is funded by a PhD grant of the Agency of Innovation  by  Science  and  Technology  in  Flanders  (IWT). We thank NVIDIA Corporation for their generous donation of a Titan X GPU.

\bibliography{submission}
\bibliographystyle{icml2017}

\end{document}